\documentclass[10pt,twocolumn,letterpaper]{article}

\usepackage[pagenumbers]{cvpr} 

\usepackage{multirow,hhline}
\usepackage{colortbl}
\usepackage{makecell}
\usepackage{float}

\newcommand\icon{\raisebox{-4pt}{\includegraphics[width=1.2em]{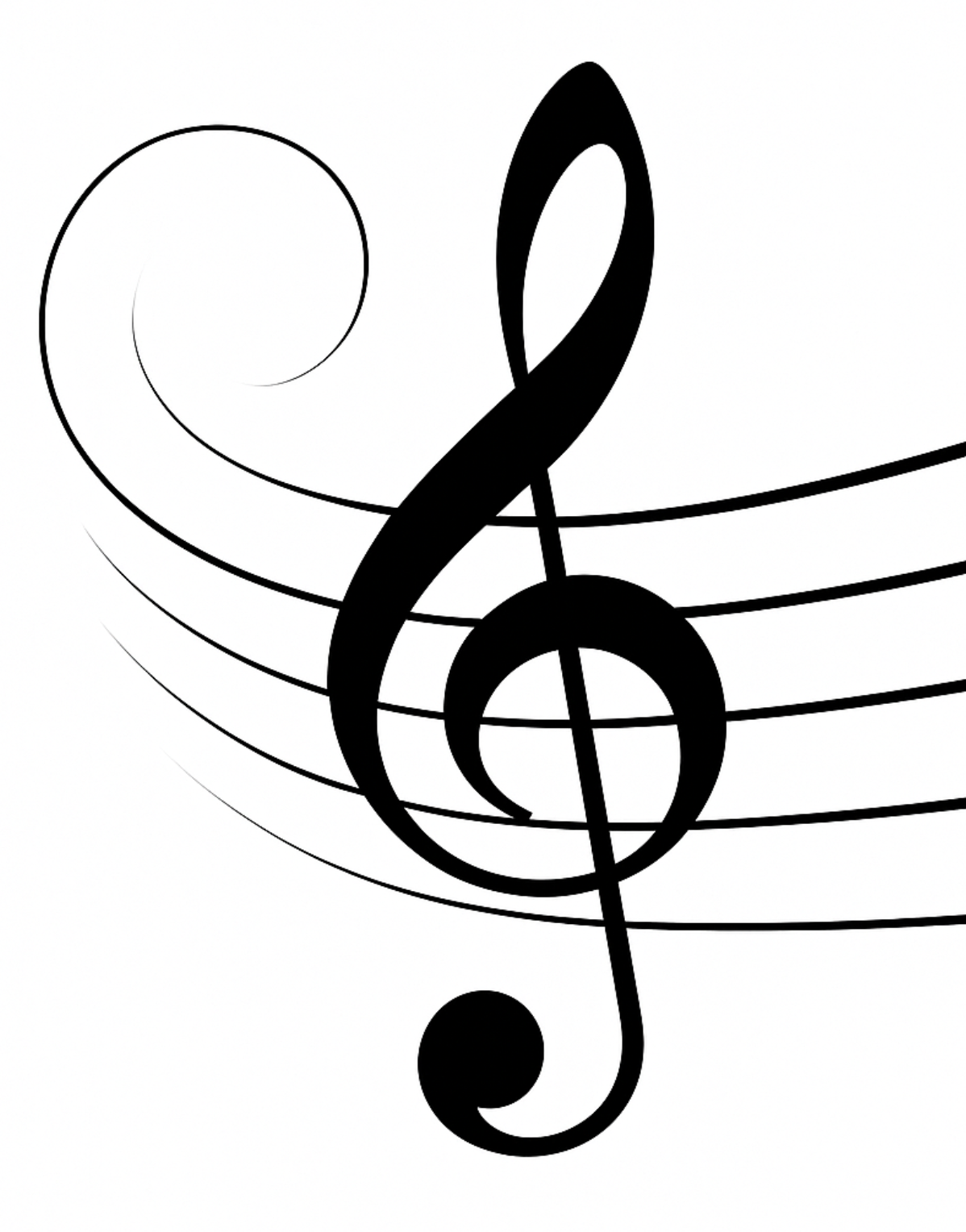}}}

\definecolor{cvprblue}{rgb}{0.21,0.49,0.74}
\usepackage[pagebackref,breaklinks,colorlinks,allcolors=cvprblue]{hyperref}

\def\paperID{7635} 
\def\confName{CVPR}
\def\confYear{2025}

\title{\fontsize{16.8}{16}\selectfont\icon VidMuse: A Simple Video-to-Music Generation Framework with Long-Short-Term Modeling}

\author{Zeyue Tian\thanks{Equal Contribution}$^{\;\,1}$, 
Zhaoyang Liu$^{*1}$, 
Ruibin Yuan$^1$, 
Jiahao Pan$^1$,\\
Qifeng Liu$^{1}$, 
Xu Tan$^2$, 
Qifeng Chen\thanks{Corresponding Authors}$\;\,^1$, 
Wei Xue$^{\dagger 1}$, 
Yike Guo$^1$\\
\\ 
\small $^{1}$Hong Kong University of Science and Technology\\
\small $^{2}$Microsoft Research Asia
}

\begin{document}
\maketitle
\begin{abstract}
In this work, we systematically study music generation conditioned solely on the video.
First, we present a large-scale dataset by collecting 360K video-music pairs, including various genres such as movie trailers, advertisements, and documentaries.
Furthermore, we propose VidMuse, a simple framework for generating music aligned with video inputs. VidMuse stands out by producing high-fidelity music that is both acoustically and semantically aligned with the video. By incorporating local and global visual cues, VidMuse enables the creation of coherent music tracks that consistently match the video content through Long-Short-Term modeling. Through extensive experiments, VidMuse outperforms existing models in terms of audio quality, diversity, and audio-visual alignment. The code and datasets are available at \url{https://vidmuse.github.io/}

\end{abstract}    
\section{Introduction}
\label{sec:intro}

Music, as an essential element of video production, can enhance humans' feelings and convey the theme of the video content. Along with the development of social media platforms \ie, YouTube and TikTok, some studies~\citep{ma2022research,dasovich2022exploring,millet2021soundtrack} have shown that a piece of melodious music can vastly attract the audience's attention and interest in watching the video. It thus leads to a great demand for studying video-to-music generation~\citep{di2021video,kang2023video2music,su2023v2meow,gan2020foley,hong2017content,he2024llms}.

Nevertheless, music creation for a video is a challenging task, which needs to understand both music theory and video semantics. It would be very time-consuming to produce a piece of suitable music for video in a hand-crafted manner. Therefore, it is desirable when we can automatically generate high-quality music for different genres of videos.
Currently, most of works~\citep{copet2024simple,huang2018musictransformer,yang2023diffsound,forsgren6riffusion,huang2023noise2music,schneider2023mo} have made significant achievements, especially in text-to-music generation, but the video-to-music generation still remains to be further studied. 
Specifically, existing works on video-conditioned music generation mainly focus on specific scenarios, such as dance videos~\citep{li2021ai,zhu2022quantized}, or on the symbolic music, \ie, MIDI~\citep{wang2020pop909,di2021video,zhuo2023video,kang2023video2music}.
However, these works are unable to generate more diverse musical styles and are also difficult to generalize to various video genres.
Moreover, Hong~\etal~\citep{hong2017content} build a music–video retrieval dataset from YouTube-8M~\citep{abu2016youtube}, albeit with limited video genres.
Despite that there are also some prominent works~\citep{hussain2023m,su2023v2meow} employing multi-modal inputs to generate music for the video, it is still worth studying that \textbf{conditioned solely on the visual input, whether it is possible to generate diverse and harmonious music for various genres of videos.}

\begin{figure*}[t!]
\centering
\includegraphics[width=0.95\linewidth]{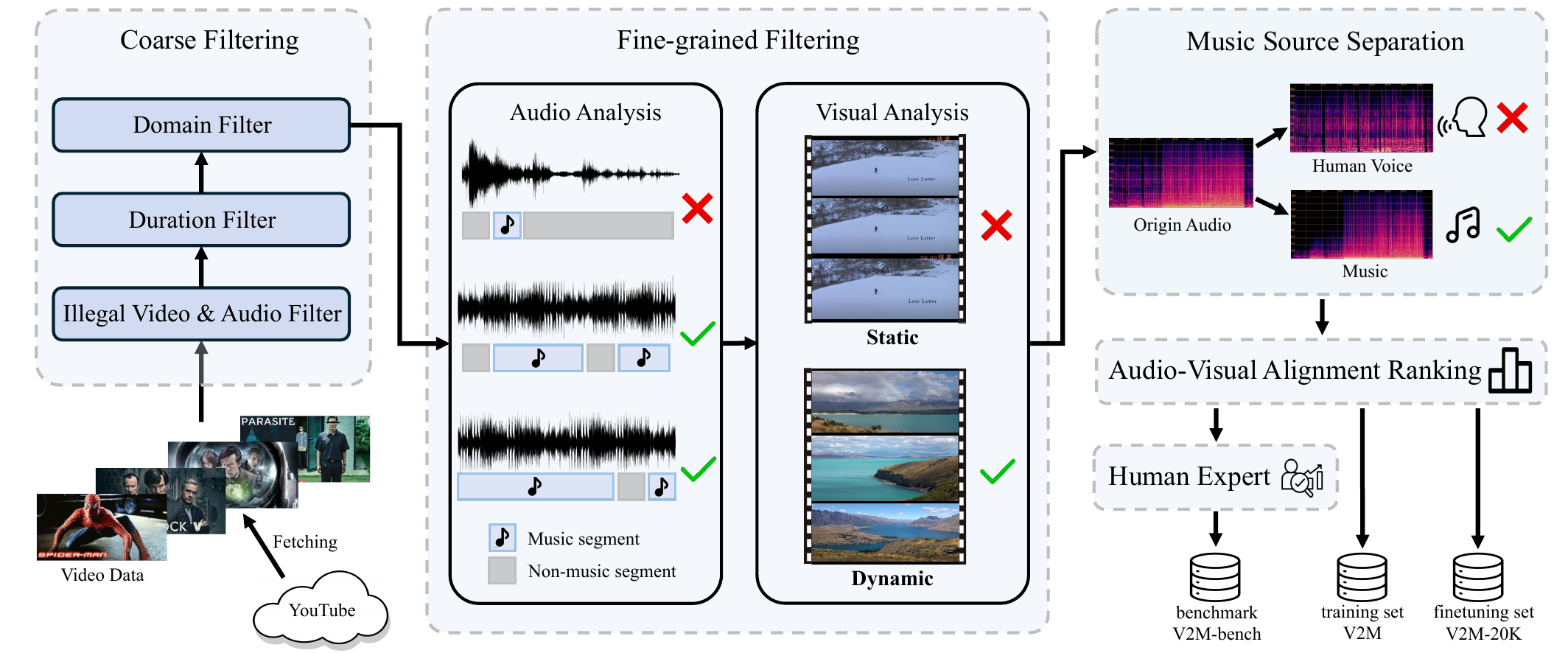}
\caption{\textbf{Dataset Construction.} To ensure data quality, \textit{V2M} goes through rule-based coarse filtering and content-based fine-grained filtering. Music source separation is applied to remove speech and singing signals in the audio. After processing, human experts curate the benchmark subset, while the remaining data is used as the pretraining dataset. The pretrain data is then refined using Audio-Visual Alignment Ranking to select the finetuning dataset.}
\label{fig:data_pipeline}
\end{figure*}

Motivated by this, we first construct a large-scale dataset termed \textit{V2M}, equipped with a comprehensive benchmark to evaluate the state-of-the-art works thoroughly. The video-music pairs are collected from YouTube with various genres, \eg, movie trailers, advertisements, documentaries, vlogs, \etc. In order to ensure the quality of our dataset, we establish a multi-step pipeline illustrated in Fig.~\ref{fig:data_pipeline} to systematically clean and pre-process data. The videos with low quality or composed of static images are filtered out. The proposed dataset contains three subsets: \textit{V2M-360K} for pretraining, \textit{V2M-20K} for finetuning, and \textit{V2M-bench} for evaluation. We believe that \textit{V2M} is able to facilitate the development of video-to-music generation. 

Furthermore, on top of \textit{V2M}, we propose a simple yet effective method, termed as \textit{VidMuse}, to generate music only conditioned on the visual input. Instead of predicting the intermediate musical symbols such as MIDI or retrieving the music from the database, the proposed VidMuse, integrates both local and global visual cues to generate background music consistent with the video in an end-to-end manner. The core techniques in our method are a \textit{Long-Short-Term Visual Module (LSTV-Module)} and a \textit{Music Token Decoder}. Specifically, the LSTV-Module aims to learn the spatial-temporal representation of videos, which is the key to generating music aligned with the video. On the one hand, the long-term module models the entire video, capturing global context to understand the whole video. It contributes to the coherence of generated music at the video level.
On the other hand, the short-term module focuses on learning the fine-grained cues at the clip level, which plays a vital role in generating diverse music.
The integration of two modules can improve the quality and visual consistency of generated music. In addition, the Transformer-based music token decoder is an autoregressive model, converting video embeddings obtained by LSTV-Module into music tokens. We formulate music generation as a task of next token prediction, which has been widely validated by the NLP community. The predicted music tokens are further decoded into the music signals by a high-fidelity neural audio compression model.

The main contributions of this work are as follows:

\begin{itemize}
\item We construct a large-scale video-to-music dataset, \ie, \textit{V2M}, which contains about 360k video-music pairs with high quality, covering various genres and themes. To the best of our knowledge, this is the largest and most diverse dataset for this task, which can facilitate future research.

\item We propose a simple yet effective method, VidMuse, for video-to-music generation. The proposed method integrates both local and global cues in the video, enabling the generation of high-fidelity music tracks that are not only musically coherent but also semantically aligned with the video content.

\item We benchmark several state-of-the-art works against our method on the V2M-bench via a series of subjective and objective metrics for a thorough evaluation. As demonstrated in experiments, VidMuse achieves state-of-the-art performance on \textit{V2M-bench}, outperforming existing models in terms of audio quality, diversity, and audio-visual consistency.
\end{itemize}
\section{Related Work}
\label{sec:related_works}

We review the existing works related to video-to-music generation, which mainly fall into four categories:

\textbf{Video Representation.}
Various methods have been proposed to learn the spatio-temporal representation ~\cite{liu2020teinet,liu2021tam,tran2018closer,feichtenhofer2019slowfast,arnab2021vivit,tong2022videomae,zhang2023video,ma2024followyourpose} for videos. They aim to capture the contextual features of video frames, which is beneficial for video understanding.
Recent advances primarily concentrate on developing video transformers~\cite{arnab2021vivit,tong2022videomae,liu2022videoswin,ma2023magicstick}. These transformer-based methods achieve superior generalized performance on various video understanding tasks, such as video classification and temporal action localization. Among them, Tong \etal~\cite{tong2022videomae} extend masked autoencoders~\cite{he2022masked} from the image to the video, exhibiting the strong generalized performance in downstream tasks. 
Benefiting from the advance in multi-modal large language models, lots of works~\cite{zhang2023video,lin2023video,munasinghe2023pg} of interactive video understanding have been proposed, which built upon the large language models (LLMs)~\cite{touvron2023llama,zheng2024judging} and showcase the visual reasoning capabilities for video understanding. 

\textbf{Audio-Visual Alignment.} Audio-visual alignment~\cite{akbari2021vatt,rouditchenko2020avlnet,shi2022learning,cheng2022joint,gong2022contrastive,wu2023next,xing2024seeing} aims to align the feature between audio, vision in the semantics level, which plays a vital role in tasks of audio-visual understanding and generation. For example, CAV-MAE~\cite{gong2022contrastive} is an audio-visual MAE that integrates the contrastive learning and masked modeling method. Currently, many works go beyond exploring audio-visual alignment. 
ImageBind~\cite{girdhar2023imagebind} extends CLIP~\cite{radford2021learning} to more modalities, including audio, depth, thermal, and IMU data, which paves the way for cross-modal retrieval and generation. In addition, Wu~\etal~\cite{wu2023next} employ LLMs with multi-modal adaptors to support any modal data as input and output, showing strong capabilities in universal multi-modal understanding. 
These methods transcend audio-visual alignment and dramatically advance the development of multi-modal representation learning.

\textbf{Conditional Music Generation.} 
Despite that there are lots of methods~\cite{dong2018musegan,hawthorne2018enabling,huang2018musictransformer,liu2020unconditional,mittal2021symbolic,lv2023getmusic,maina2023msanii} studying unconditional music generation, in this paper, we mainly focus on reviewing the methods of conditional music generation, which are more related to our work.
Many researchers~\cite{schneider2023mo,agostinelli2023musiclm,yang2023diffsound,forsgren6riffusion,huang2023noise2music,copet2024simple,yuan2024chatmusician, deng2024composerx} make their endeavours on text-to-music generation. Similar to Stable Diffusion~\cite{rombach2022high}, these works~\cite{yang2023diffsound,forsgren6riffusion,huang2023noise2music,schneider2023mo} try to adapt diffusion models for music generation.
M$^2$UGen~\cite{hussain2023m} is a multi-modal music understanding and generation system that leverages large language models to process video, audio, and text. 
Video2Music\cite{kang2023video2music} can generate music that matches the content and emotion of a given video.
Moreover, the proposed V2Meow~\cite{su2023v2meow} and MeLFusion~\cite{chowdhury2024melfusion} conditioned on video and image, respectively, can generate music that supports style control via text prompts.
In contrast to previous video-to-music works~\cite{gan2020foley,kang2023video2music,zhuo2023video,su2023v2meow}, our VidMuse utilizes a short-term module and a long-term module to model local and global visual cues in videos. As a result, it can generate high-fidelity music aligned with the video.

\begin{figure*}[tb]
  \centering
  \begin{subfigure}{0.52\linewidth} 
    \includegraphics[width=1\textwidth]{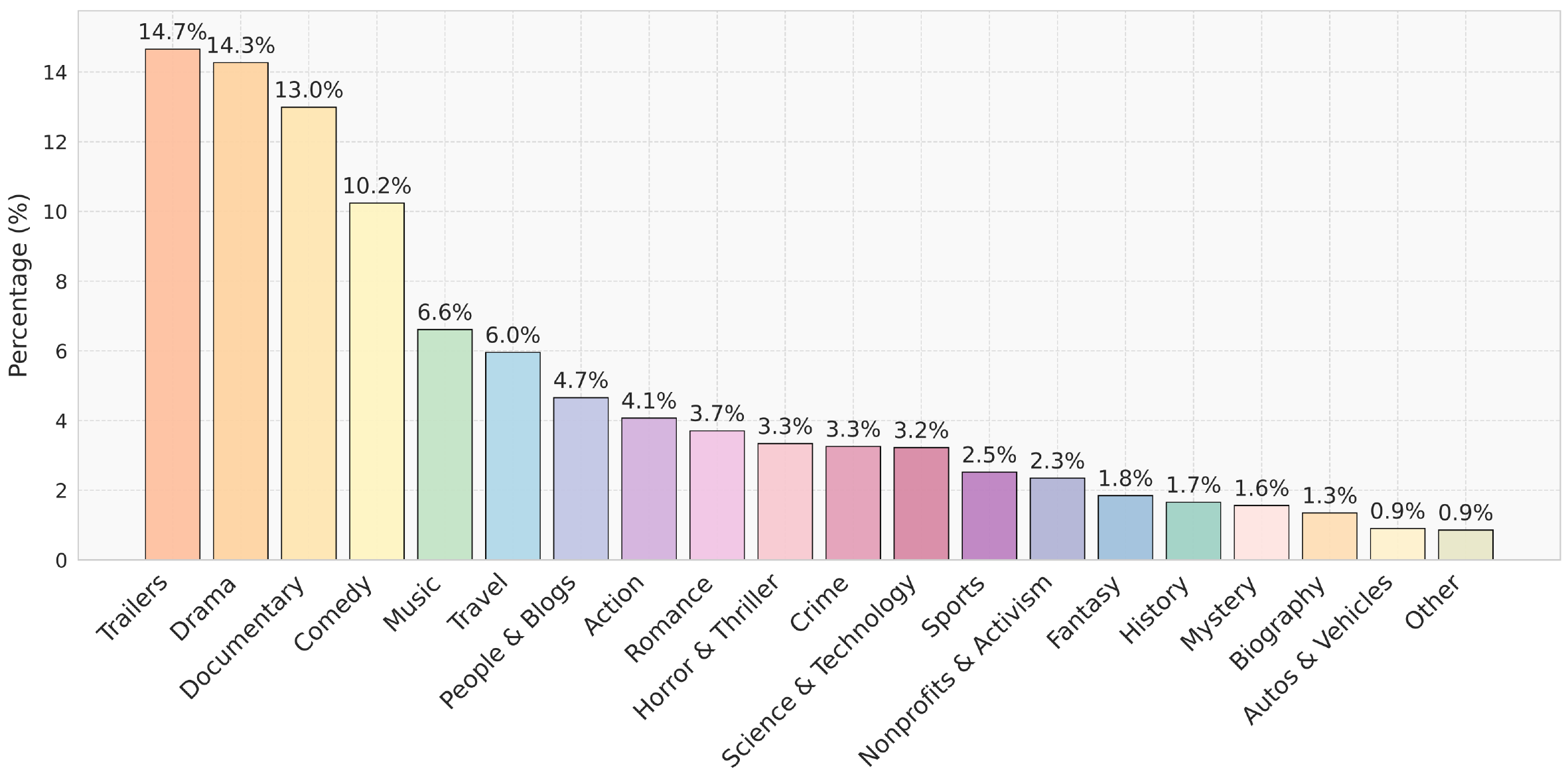}
    \caption{Video genre distribution.}
    \label{fig:distribution_genres}
  \end{subfigure}
  \hspace{0.05\linewidth}
  \begin{subfigure}{0.27\linewidth}
    \includegraphics[width=1\textwidth]{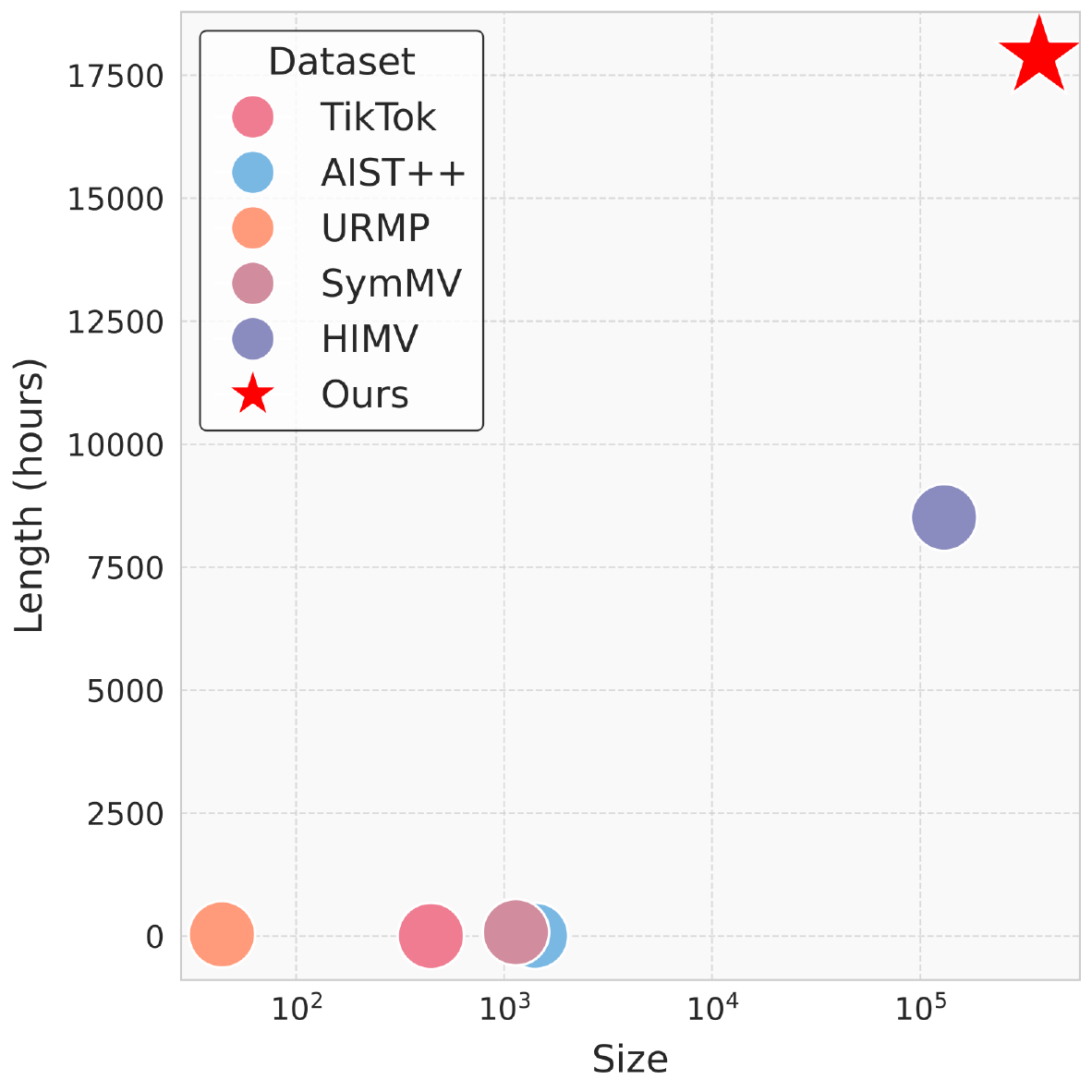}
    \caption{Dataset Comparisons.}
    \label{fig:Comparisons_datasets}
  \end{subfigure}
\caption{\textbf{Statistics of our dataset.} (a) The distribution of video genres in our dataset, (b) Comparisons with other related datasets in terms of scale of datasets. Please zoom in for details.}
  \label{fig:short}
\end{figure*}

\textbf{Video-to-music Datasets.} Many multimodal datasets~\cite{changpinyo2021conceptual,tian2020unified,lee2021acav100m,gemmeke2017audio,zhou2018towards,miech2019howto100m,srinivasan2021wit,abu2016youtube,schuhmann2022laion,hershey2017cnn,xue2022hdvila,chen2024panda, ma2024foundation} have been released, but there is still a lack of datasets for video-to-music generation. Hong~\etal~\cite{hong2017content} construct the HIMV-200K with video-music pairs and aim to retrieve music for the video from the database. However, this dataset exhibits limited video genres and also suffers from the issue of data quality, as stated in \cite{zhuo2023video}.
We observe that several works~\cite{wang2020pop909,di2021video,zhuo2023video,li2024diff} aim to facilitate MIDI music generation. However, this musical form imposes restrictions on diversity for the music generation. Other datasets~\cite{zhu2022quantized,li2021ai} focus on generating music for dance videos only and have limited data size, which limits their applicability for general video-to-music models. As a result, this work constructs a large-scale video-to-music dataset where the music directly in wav format is diverse. We establish a rigorous pipeline for data collection and cleaning, which ensures the quality and diversity of our dataset. We expect the model can learn the music with more diverse forms from this dataset.

\section{Dataset}
In this section, we build a multi-step pipeline to clean and process source videos from YouTube to ensure data quality. After that, we construct a large-scale video-to-music generation dataset, \ie, \textit{V2M}, with a benchmark. 
The constructed dataset stands out for its significant size, high quality, and rich diversity, including a wide range of genres such as movie trailers, advertisements, documentaries, vlogs, \etc. This comprehensive and diverse dataset aims to facilitate the video-to-music generation.

\subsection{Dataset Collection}

To quickly collect a large scale of video-music pairs, we curate a series of query sets to retrieve corresponding videos from YouTube. In addition, we find that the music in the movie trailer usually showcases rich diversity and high quality. Therefore, we also aggregate a vast array of video information from the IMDb Non-Commercial Datasets, including video types, names, release dates, etc. Queries are formulated based on the titles of these selections and retain the videos released after 2000, as videos from earlier periods are less likely to be of good quality. In the process of video crawling, we only keep the top 2 search results, resulting in a collection of around 400K videos, ranging from movies to documentaries.
Besides, several existing datasets already contain video-music pairs, such as HIMV-200K~\cite{hong2017content}, subsets of YouTube-8M~\cite{abu2016youtube} labeled with ``Music'' and ``Trailer'' tags. We incorporate these datasets into our collection to further expand its scope. After merging all sources, our final dataset comprises about 600K videos, spanning diverse genres and categories.

\subsection{Dataset Construction}
The raw videos may include many low-quality samples. To address this, we develop a series of rigorous steps to filter out undesirable data and obtain a clean set. The overall pipeline of data processing is depicted in Fig. \ref{fig:data_pipeline}.
The following steps outline our approach:
(1) The process begins with coarse filtering, where we remove videos lacking audio or video tracks, videos that are too short or too long, those containing inappropriate content such as violence or explicit material, and those from categories like \textit{Interview} and \textit{News}, which generally have background music not aligned with the visual content.
(2) Following that, we perform fine-grained filtering to retain videos with substantial music content and dynamic visual elements. We use an audio analysis model~\cite{kong2020panns} to identify music segments, ensuring a sufficient portion of the audio is classified as music. In parallel, we analyze the visuals~\cite{wang2004image} to exclude videos consisting mainly of static images.
(3) To further refine the dataset, we apply music source separation~\cite{rouard2023hybrid} to isolate the music component by removing vocal tracks, enhancing the overall audio quality.
(4) Finally, we rank the videos based on their audio-visual alignment scores~\cite{girdhar2023imagebind} to ensure a high level of semantic correlation between the audio and visual modalities.
The resulting videos are then split into training (\textit{V2M}), fine-tuning (\textit{V2M-20K}), and benchmark (\textit{V2M-bench}) subsets. For details on dataset construction, please refer to the Appendix~\ref{sec:Details_of_Dataset_Construction}).

\subsection{Data Analysis}
The above data pipeline yields three data splits. Specifically, the training set comprises $\sim$360K video-music pairs, around $1.8\times 10^4$ hours. The finetuning dataset consists of $\sim$20K pairs, about $6 \times 10^2$ hours. The benchmark dataset contains 300 pairs, with a cumulative duration of 9 hours.
Fig.~\ref{fig:distribution_genres} showcases the genre
distribution of our training data, highlighting its comprehensive diversity. This diversity ensures a rich and varied dataset for the model training. As shown in Fig.~\ref{fig:Comparisons_datasets}, we compare with other related datasets, demonstrating its advantage in data scale. 

\noindent\textbf{Dataset Necessity.} Some existing video-music pair datasets have been released~\cite{hong2017content, wang2020pop909,di2021video,zhuo2023video, zhu2022quantized,li2021ai}, but
some of them~\cite{wang2020pop909,di2021video,zhuo2023video} aim to
facilitate MIDI music generation, which limits the form of music. Datasets like~\cite{zhu2022quantized,li2021ai} focus on generating music for dance videos only and have limited data size. The dataset constructed by~\cite{hong2017content} includes video-music pairs but exhibits limited video genres and suffers from data quality issues. In addition, the evaluation metrics used in these video-to-music benchmarks~\cite{wang2020pop909,hong2017content,di2021video,yu2023long,li2024diff,li2021ai,zhu2022quantized,zhuo2023video} are divergent, making it difficult to fairly and thoroughly assess performance of methods on different benchmarks. Motivated by this, we develop the multi-step pipeline and curate a large-scale dataset \textit{V2M} as well as a benchmark for the video-to-music generation.

\section{Method}

\begin{figure*}[t]
\centering
\includegraphics[width=1\linewidth]{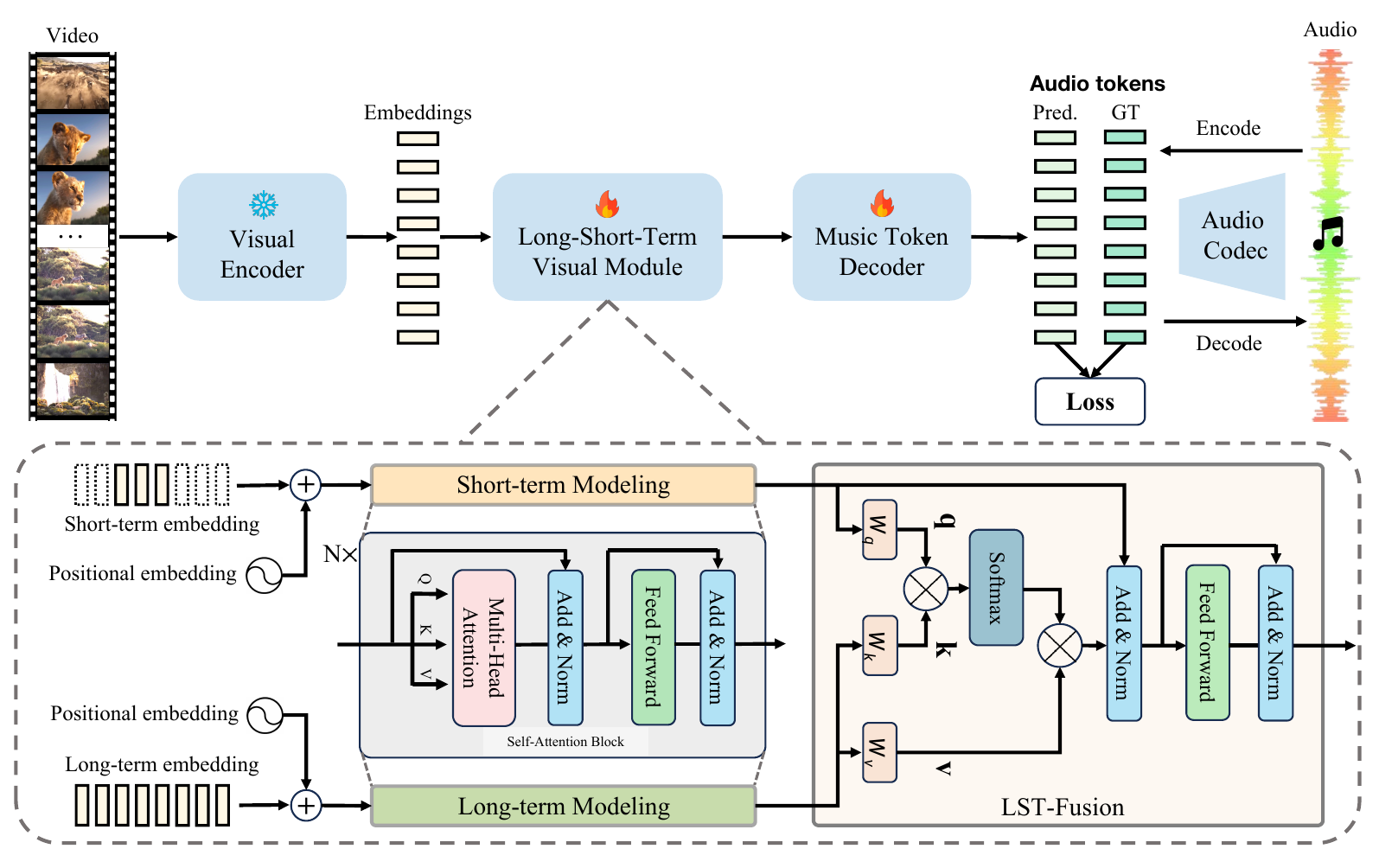}
\caption{\textbf{Overview of the VidMuse Framework.} This pipeline outlines the key components for generating music aligned with video content: (1) Visual Encoder for extracting visual features, (2) Long-Short-Term Visual Module for integrating local and global cues, (3) Music Token Decoder for generating music tokens, and (4) Audio Codec for the conversion between audio and audio tokens.}
\label{fig:pipeline}
\end{figure*}

\subsection{Architecture of VidMuse}
In this section, we elaborate on the framework of VidMuse, which leverages \textbf{LSTV-Module} to generate music aligned with video content. The proposed pipeline is shown in Fig.~\ref{fig:pipeline}, including (1) Visual Encoder, (2) LSTV-Module, (3) Music Token Decoder, and (4) Audio Codec.

\noindent\textbf{Visual Encoder.} To generate music conditioned on the video, we first need to extract the high-level features from a stack of frames. Given an input video, the visual encoder extracts feature representations $\textbf{X} \in \mathbb{R}^{N \times P \times D}$. Here, N is the number of input frames, $P$ refers to the sequence length with the class token, and $D$ denotes the size of the feature vectors. Currently, there are lots of visual encoders available, including 2D~\cite{dosovitskiy2020image}, 3D~\cite{tong2022videomae,arnab2021vivit} and multi-modal~\cite{radford2021learning} models, which will be validated in the Sec.~\ref{sec:ablation_study}.

\noindent\textbf{LSTV-Module.} 
Generating music for videos with variable length still presents significant challenges, especially for a long video, because sometimes it is difficult to directly model on whole video due to hardware limitations. Prior studies opt to generate music in segment level~\cite{kang2023video2music,di2021video,hussain2023m}. However, such a manner often lacks sufficient context information. Music should vary in expression depending on the context. Even the same video segment may lead to distinct musical interpretations when being in different contexts. By incorporating global guidance, it can enhance the alignment of the generated music with the overall video content.
To capture both local and global visual cues, the visual features extracted from the visual encoder are thus fed into the LSTV-Module. 
Specifically, the short-term module takes segment-level embeddings as input, aiming to capture local dependencies $\textbf{X}_s  \in \mathbb{R}^{N_s \times P \times D}$ to ensure that the generated music aligns with short-term variations in the video, while long-term module models on video-level embeddings, providing context $\textbf{X}_l \in \mathbb{R}^{N_l \times P \times D}$ to guide the short-term module in generating more visually coherent music. $N_s$ and $N_l$ is the number of frames sampled from the video.

To capture both global and local visual cues, we leverage the cross-attention~\cite{vaswani2017attention} in both Long-term and Short-term Modeling. Long-term modeling extracts long-range dependencies, while short-term modeling focuses on local details. This results in refined long-term features $\textbf{F}_l \in \mathbb{R}^{N_l \times P \times D}$ and short-term features $\textbf{F}_s \in \mathbb{R}^{N_s \times P \times D}$.

To incorporate global guidance for generating segment-based music, we design LST-Fusion. It integrates long-term and short-term features by utilizing the cross-attention mechanism $\text{CA}(\cdot)$ with Query ($\mathbf{Q}$), Key ($\mathbf{K}$), and Value ($\mathbf{V}$), which can be mathematically formulated as:
\begin{equation}
  \label{eq:cross_attention}
  \begin{split}
      \textbf{Z}' = \text{CA}&(\mathbf{Q}, \mathbf{K}, \mathbf{V}),\\ 
  \text{ where } \mathbf{Q} = \textbf{F}_s&, \mathbf{K} = \textbf{F}_l, \mathbf{V} = \textbf{F}_l,
  \end{split}
\end{equation}

This mechanism allows the model to query global information rather than generating music based solely on local visual features. It guarantees that the generated music is more consistent with the video content.
After the cross-attention, a linear layer projects $\textbf{Z}'$ to $\textbf{Z} \in \mathbb{R}^{N_s \times P \times M}$, where $M$ represents the input vector dimension of the music token decoder in the next step. In addition, we also explore different implementations discussed in experiments.

\begin{table*}[!t]
\caption{
\small{\textbf{Comparison with naive baselines and state-of-the-art methods.}} 
}
\label{tab:main_result}
\centering
\setlength{\tabcolsep}{1mm}
\renewcommand{\arraystretch}{1.1}
\begin{tabular}{c|c|c|c|c|c|c}
\Xhline{1.5pt}

\multicolumn{1}{c|}{\multirow{2}{*}{Methods}} &  \multicolumn{6}{c}{Metrics} \\
\cline{2-7}
& KL$~\color{red}{\downarrow}$& FD$~\color{red}{\downarrow}$ & FAD$~\color{red}{\downarrow}$ &  density$~\color[RGB]{50, 200, 50}{\uparrow}$ & coverage$~\color[RGB]{50, 200, 50}{\uparrow}$ & Imagebind$~\color[RGB]{50, 200, 50}{\uparrow}$ \\
 \Xhline{1pt}
 \rowcolor{gray!30} 
GT & $0.000$ & $0.000$ & $0.000$ & $1.167$ & $1.000$ & $0.241$ \\
Caption2Music & $1.081$ & $40.199$ & $2.485$ & $ 0.378$ & $0.486$ & $0.191$ \\
Video2Music~\citep{kang2023video2music} & $1.782$ & $144.881$ & $18.722$ & $ 0.103$ & $0.023$ & $0.136$\\
CMT~\citep{di2021video} & $1.220$ & $85.704$ & $8.637$ & $0.080$ & $0.070$ & $0.124$\\
M$^2$UGen~\citep{hussain2023m} & $0.997$ & $52.246$ & $5.104$ & $0.608$ & $0.433$  & $0.181$ \\
M$^2$UGen*~\citep{hussain2023m} & $0.965$ & $52.041$ & $5.003$ & $0.633$ & $0.430$  & $0.180$ \\
VM-NET~\citep{hong2017content} & $0.899$ & $67.480$ & $6.252$ & $0.986$ & $0.383$ & $0.147$\\ 
 \hline
VidMuse & $\textbf{0.734}$ & $\textbf{29.946}$ & $\textbf{2.459}$ & $\textbf{1.250}$ & $\textbf{0.730}$ & $\textbf{0.202}$ \\
\Xhline{1.5pt}
\end{tabular}
\end{table*}
\noindent\textbf{Music Token Decoder.} We adopt an autoregressive approach to predict the music tokens $\bar{\textbf{Y}}$ conditioned on the video segment. Music token decoder is implemented by a transformer decoder with a linear classifier. We set the latent vector size of the transformer decoder to $M$, allowing it to scale up or down the model's size. The decoder incorporates a cross-attention mechanism that receives the visual signal $\textbf{Z} \in \mathbb{R}^{N_s \times P \times M}$, where $N_s$ is the number of frames sampled in the video segment.
At each time step $t$ (where $t = 1, \dots, T$), the decoder predicts the logits of current token $\bar{\textbf{Y}}_{t} \in \mathbb{R}^{K \times C}$ based on previous tokens and visual context $\textbf{Z}$. Here, $K$ denotes the number of codebooks, and $C$ represents the vocabulary size.

\noindent\textbf{Audio Codec.} 
It can convert an audio segment into discretized codebooks and, conversely, decode codebooks back into audio. The size of codebooks is $K \times T$, where $T$ denotes the length of the video. Given the Audio codec $\mathcal{C}$, we denote the Encoder as $\mathcal{C}_{\text{encode}}(\cdot)$ and the Decoder as $\mathcal{C}_{\text{decode}}(\cdot)$. In training, we need to encode the ground truth audio $\textbf{A}$ into discretized tokens serving as supervise signals for the next token prediction. In the inference phase, the predicted tokens will be then decoded into music signals.

\subsection{Training}

Given a video segment with corresponding ground-truth audio $\textbf{A}$, we train our model using a next-token prediction approach. The video segments are processed through the Visual Encoder and the LSTV-Module to generate visual features, which are then fed into the Music Token Decoder to produce the predicted logits $\bar{\textbf{Y}} \in \mathbb{R}^{K \times T \times C}$. Here, $K$ denotes the number of codebooks, $T$ is the sequence length (number of timesteps), and $C$ is the vocabulary size for codebooks.
Next, the ground-truth audio $\textbf{A}$ is encoded by the Audio Codec to obtain the target one-hot vector: $\textbf{Y} = \mathcal{C}_{\text{encode}}(\textbf{A})$, where $\textbf{Y} \in \mathbb{R}^{K \times T \times C}$.
The value of $\textbf{Y}_{k,t,c}$ is 1 when $c$ equals the ground-truth token index in codebook $k$ at timestep $t$, and 0 otherwise.

Our objective is to minimize the cross-entropy loss between the predicted probabilities $\bar{\textbf{Y}}$ and the ground-truth tokens $\textbf{Y}$. The cross-entropy loss $\mathcal{L}$ is defined as:

\begin{equation}
\label{eq:cross_entropy_loss}
\mathcal{L} = -\frac{1}{K T} \sum_{k=1}^{K} \sum_{t=1}^{T} \sum_{c=1}^{C} \textbf{Y}_{k,t,c} \log \bar{\textbf{Y}}_{k,t,c},
\end{equation}

where $\bar{\textbf{Y}}_{k,t,c}$ is the predicted probability of class $c$ at codebook $k$ and timestep $t$.
Conditioned on the video, we train the model by predicting the next token with this loss.

\section{Experiments}
\label{sec:Experiments}

In this section, we elaborate on the implementation details of our experiments, and conduct massive experiments to thoroughly evaluate the efficacy of our proposed method from both subjective and objective perspectives. This is expected to provide insights for video-to-music generation.

\subsection{Implementation details}

Since this work does not focus on audio encoding and decoding, we use Encodec~\cite{defossez2022highfi} for 32 kHz monophonic audio as our default compression model and use the pretrained transformer model proposed in MusicGen~\cite{copet2024simple}. 
The training stage utilizes the AdamW optimizer~\cite{loshchilov2017decoupled} with a batch size of 5 samples per GPU. We sample frames from a continuous 30s video segment at 2 fps for short-term modeling and uniformly sample 32 frames from the entire video for long-term modeling. The hyperparameters are set to $\beta_1=0.9$, $\beta_2=0.95$, with a weight decay of 0.1 and gradient clipping at 1.0. A cosine learning rate schedule is employed, incorporating a warm-up phase of 4,000 steps and an exponential moving average decay of 0.99. 
We use 64 H800 GPUs in pretraining and train models with $56K$ steps, which takes about 50 hours.
For the finetuning stage, we utilize 32 H800 GPUs and train models with $8K$ steps, which takes about 8 hours.
A top-k strategy is applied for sampling, retaining the top $250$ tokens with a temperature setting of $1.0$. In the inference stage, we set the sliding window size as $30$s, and the window's overlap as $0.5$s.

\subsection{Evaluation Metrics}
To quantitatively evaluate the effectiveness of our model, we employ a series of metrics to assess different models in terms of quality, fidelity, and diversity of the generated music. These metrics include the Frechet Audio Distance (FAD), Frechet Distance (FD), Kullback-Leibler Divergence (KL), as well as Density and Coverage~\cite{naeem2020reliable}. Additionally, we utilize the ImageBind Score~\cite{girdhar2023imagebind} to examine the alignment between the video and the generated music. We acknowledge that ImageBind has limitations as it is not specifically trained on music data, but it currently seems to be a possible option for evaluating the semantic alignment between video and generated music. 
To fairly compare with non-public baseline models~\cite{yu2023long, zhu2022quantized, su2023v2meow}, we use evaluation metrics from their papers: beats coverage score (BCS), beats hit score (BHS), standard deviations of BCS (CSD) and BHS (HSD), and F1 scores of BCS and BHS. 


\subsection{Main Results}

We benchmark several state-of-the-art methods, serving as baselines to compare with our method: 
1) \textbf{Caption2Music}, a naive baseline that employs the SpaceTimeGPT to extract the video captions and outputs the music by feeding captions into MusicGen~\cite{copet2024simple}.
2) \textbf{Video2Music}~\cite{kang2023video2music} and 3) \textbf{CMT}~\cite{di2021video} which both predict MIDI notes~\cite{rothstein1995midi} from videos while our method directly generates music signals. 
4) \textbf{M$^2$UGen}~\cite{hussain2023m}, a strong baseline, which leverages a language model to connect vision and language, then use MusicGen~\cite{copet2024simple} to generate music from language.
5) \textbf{M$^2$UGen*}, a re-trained version of M$^2$UGen using our dataset.
6) \textbf{VM-NET}~\cite{hong2017content}, different from above methods, retrieves a piece of music from the database for a given video, while other methods predict music by training on video-music pairs. 

In Table~\ref{tab:main_result}, VidMuse, with both global and local visual modeling, exhibits impressive performance on all metrics. Specifically, compared with Video2Music~\cite{kang2023video2music} and CMT~\cite{di2021video}, VidMuse shows the superiority in the diversity of generated music based on the density or coverage. It justifies the advantage of directly predicting music signals compared with MIDI notes. 
Our method even outperforms the strong competitors, \ie, M$^2$UGen. It proves that our method of directly predicting music based on video input can also achieve better performance. 
Furthermore, compared with a retrieval-based method, \ie, VM-NET~\cite{hong2017content}, VidMuse achieves a higher Imagebind score, indicating that the music generated by the learning-based strategy is more consistent with the video semantics.

\begin{table}[!t]
\caption{
    \small{\textbf{Results across different benchmarks~\cite{yu2023long,zhu2022quantized,li2021ai}.}}
    }
    \centering
    \label{tab:results_on_different_benchmarks}
    \footnotesize
    \renewcommand{\arraystretch}{1.2}
    \begin{tabular}{c|c|c|c|c|c}
    \Xhline{1.5pt}
     \multicolumn{1}{c|}{\multirow{2}{*}{Methods}} & \multicolumn{5}{c}{Metrics} \\
    \cline{2-6}
    & BCS$~\color[RGB]{50, 200, 50}{\uparrow}$ & CSD$~\color{red}{\downarrow}$ & BHS$~\color[RGB]{50, 200, 50}{\uparrow}$ & HSD$~\color{red}{\downarrow}$ & F1$~\color[RGB]{50, 200, 50}{\uparrow}$ \\
    \Xhline{1pt}

    \rowcolor{gray!15} \multicolumn{6}{c}{AIST++~\cite{li2021ai}}\\
    \hline 
    D2M-GAN~\cite{zhu2022quantized} & $92.3$ & $-$ & $91.7$ & $-$ & $-$\\        
    CDCD~\cite{zhu2022discrete} & $93.9$ & $1.2$ & $90.7$ & $\textbf{1.5}$ & $-$ \\
    V2Meow~\cite{su2023v2meow} & $\textbf{100.0}$ & $\textbf{0.0}$ & $84.4$ & $25.1$ & $91.5$\\
    VidMuse & $99.97$ & $0.3$& $\textbf{96.5}$ & $9.4$ & $\textbf{98.2}$ \\
    \hline 
    \rowcolor{gray!15} \multicolumn{6}{c}{LORIS~\cite{yu2023long}}\\
    \hline
    D2M-GAN~\cite{zhu2022quantized} & $95.6$ & $9.4$ & $88.7$ & $19.0$ &  $93.1$\\     
    CDCD~\cite{zhu2022discrete} & $96.5$ & $9.1$ & $89.3$ & $18.1$ & $92.7$ \\   
    LORIS\cite{yu2023long} & $\textbf{98.6}$ & $\textbf{6.1}$ & $90.8$ & $13.9$ & $94.5$\\
    VidMuse & $96.3$ & $10.6$ & $\textbf{95.6}$ & $\textbf{8.9}$& $\textbf{95.9}$\\
    \hline
    \rowcolor{gray!15} \multicolumn{6}{c}{TikTok~\cite{zhu2022quantized}}\\
    \hline 
    
    D2M-GAN~\cite{zhu2022quantized} & $87.1$ & $-$ & $83.9$ & $-$ &  $-$\\           
    CDCD~\cite{zhu2022discrete} & $\textbf{91.8}$ & $-$ & $86.3$ & $-$ & $-$ \\
    VidMuse & $79.8$& $20.0$ & $\textbf{97.3}$ & $7.6$ &  $87.7$ \\
    \Xhline{1.5pt}

    \end{tabular}
\end{table}

Furthermore, we validate the generalization abilities of our proposed VidMuse on several different types of video-to-music benchmarks. To make fair comparisons and avoid data leakage, we check our dataset and remove the potential repeated data sample from our training set. As shown in Table~\ref{tab:results_on_different_benchmarks}, our VidMuse achieves comparable performance against other methods.
This demonstrates that VidMuse does not overfit on our own benchmarks and exhibits strong generalization capabilities.

\begin{table}[!t]
\caption{
\small{\textbf{Ablation studies on design choices.}} 
}
\label{tab:ablation_design_choice}
\centering
\footnotesize
\renewcommand{\arraystretch}{1.2}
\begin{tabular}{c|c|c|c|c}
\Xhline{1.5pt}

Methods & KL$~\color{red}{\downarrow}$& FD$~\color{red}{\downarrow}$ & FAD$~\color{red}{\downarrow}$ &  density$~\color[RGB]{50, 200, 50}{\uparrow}$ \\
\Xhline{1pt}
VidMuse-STM& $0.898$ & $45.752$ & $4.915$ & $1.124$ \\
VidMuse-LTM& $0.858$ & $53.907$ & $16.074$ & $\textbf{1.439}$ \\
\hline
VidMuse-CAQ\_SL & $0.843$ & $48.940$ & $3.733$ & $0.947$ \\
VidMuse-CAQ\_LS & $0.919$ & $45.335$ & $2.917$ & $0.562$ \\
VidMuse-Slowfast & $1.511$ & $84.683$ & $10.029$ & $0.266$ \\
\hline
VidMuse & $\textbf{0.738}$ & $\textbf{36.171}$ & $\textbf{2.369}$ & $1.175$ \\ 
\Xhline{1.5pt}
\end{tabular}
\end{table}

\begin{table}[!t]
\caption{
    \small{\textbf{Ablation studies on visual encoders.}} 
    }
    \centering
    \label{tab:ablation_encoder}
    \footnotesize
    \renewcommand{\arraystretch}{1.1}
    \begin{tabular}{>{\centering\arraybackslash}p{1.15cm}|>{\centering\arraybackslash}p{0.57cm}|>{\centering\arraybackslash}p{0.82cm}|>{\centering\arraybackslash}p{1.02cm}|>{\centering\arraybackslash}p{1cm}|>{\centering\arraybackslash}p{1.3cm}}
    \Xhline{1.5pt}
    Encoders & KL$~\color{red}{\downarrow}$ & density$~\color[RGB]{50, 200, 50}{\uparrow}$ & GFLOPs$~\color{red}{\downarrow}$ & Latency$~\color{red}{\downarrow}$ & Throughput$~\color[RGB]{50, 200, 50}{\uparrow}$ \\
    \Xhline{1pt}
    ViViT & $0.822$ & $\textbf{1.433}$ & $451.83$ & $1650$ ms & $9.12$\\
    VideoMAE & $0.778$ & $1.074$ & $360.99$ & $452$ ms & $17.44$ \\
    CLIP & $\textbf{0.753}$ & $1.122$ & $\textbf{141.24}$ & $\textbf{341}$ ms & $\textbf{24.16}$ \\     
    ViT & $0.876$ & $1.081$ & $562.64$ & $405$ ms & $23.84$ \\
    \Xhline{1.5pt}
    \end{tabular}
\end{table} 

\begin{figure*}[!t]
\centering
\includegraphics[width=1\linewidth]{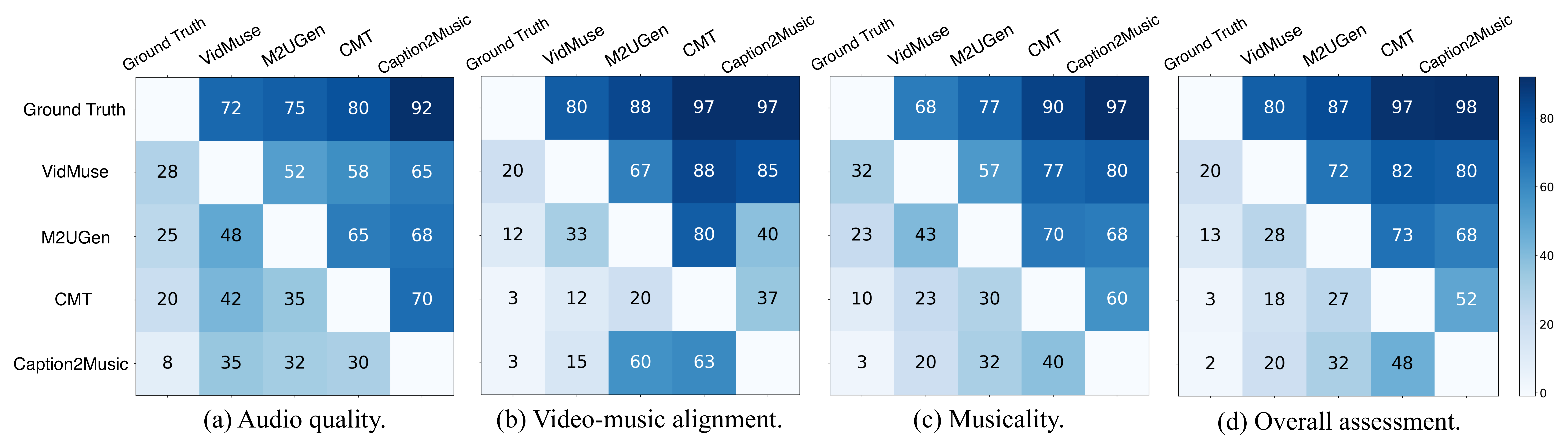}
  \caption{
    \textbf{A/B test results of the user study.} We design four criteria in Sec.~\ref{sec:user_study} to assess the subjective performance.
  }
\label{fig:ab_test_sup}
\end{figure*}

\subsection{Subjective Comparisons with User Study}
\label{sec:user_study}
In the user study, we randomly sample 600 video-music pairs from the benchmark to conduct an A/B test, which is a widely used subjective evaluation method in the music field~\cite{donahue2023singsong, yuan2024chatmusician}. This test includes CMT, M$^2$UGen, Caption2Music, Ground Truth (GT), and VidMuse. The test was distributed among 40 participants, ensuring each method was compared against another 60 times. The evaluation criteria are four-fold:
1) \textbf{Audio quality}: Refers to the sound quality of the audio track; 2) \textbf{Video-music alignment}: Assesses how well the music matches the visual content, e.g., a scene showing a woman crying should ideally be paired with music that sounds sad; 3) \textbf{Musicality}: Evaluates the aesthetic quality of the music, distinct from audio quality. For example, a piece of music may have good audio quality, but if it is out of tune, it would be considered to have poor musicality;
4) \textbf{Overall assessment}: Comprehensively evaluates the performance for models.
Participants are asked to choose the better sample for each criterion. 
The user study is shown in Fig.~\ref{fig:ab_test_sup}, where the value at matrix[$i$][$j$] ranges from $0$ to $100$, indicating the \% of times listeners preferred the method in $i$-row compared to the method in $j$-column.
For example, in Fig.~\ref{fig:ab_test_sup} (c), the value of matrix[$2$][$4$] represents that VidMuse outperforms CMT in $77\%$ of cases in terms of Musicality. Across all criteria, our method surpasses others in more than half of the comparisons, except when compared to the ground truth. Overall, these results thoroughly validate VidMuse's effectiveness through subjective evaluation.

\subsection{Ablation Studies}
\label{sec:ablation_study}
In this section, we conduct ablation studies, aiming to find the optimal design choices. 

\noindent \textbf{Justification of Design Choices.}
To validate the impact of different model design choices on our generation results and verify the effectiveness of our method, we first design two modules: a short-term modeling module (STM) and a long-term modeling module (LTM). VidMuse-STM aims to ablate the contribution of STM by removing LTM, while VidMuse-LTM utilizes only LTM. 
Based on the results in Table~\ref{tab:ablation_design_choice}, we gain the insight that local information plays a more important role in the generation. By integrating global guidance with local information, we improve the alignment of the generated music with the video content. Second, we implement two variants with \textbf{C}ross-\textbf{A}ttention with learnable \textbf{Q}ueries (\textbf{CAQ}) in our framework. Specifically, CAQ\_SL first uses a \textbf{CAQ} where $K$ and $V$ are short-term features and then uses a \textbf{CAQ} where $K$ and $V$ are long-term features. CAQ\_LS does it in the opposite order.
As shown in Table~\ref{tab:ablation_design_choice}, VidMuse outperforms two variants, demonstrating the efficacy of our manner.
Furthermore, we evaluate a baseline that replaces the LSTV-Module with a SlowFast-like mechanism~\cite{feichtenhofer2019slowfast}, where the slow path models appearance and the fast path captures temporal dynamics. As shown in Table~\ref{tab:ablation_design_choice}, this modification degrades performance, likely because the fast path operates at a higher frame rate while sharing the same temporal receptive field as the slow path, resulting in insufficient global guidance.

\noindent\textbf{Visual Encoder.} 
\label{ablate:visual_encoders}
We here study the impact of various visual encoders. As shown in Table~\ref{tab:ablation_encoder}, we experiment with different visual encoders, including ViT~\cite{dosovitskiy2020image}, CLIP~\cite{radford2021learning}, VideoMAE~\cite{tong2022videomae}, and ViViT~\cite{arnab2021vivit}. For fair comparisons, these encoders all use ViT-B as the backbone. 
The Latency and throughput are assessed with 30-second videos on the NVIDIA H800 GPU. 
Latency is measured with a batch size of 1, while throughput is measured with a batch size of 16. Our results show that VidMuse remains robust in processing visual information for music generation across all encoder choices. To balance computational efficiency and generation quality, we select CLIP~\cite{radford2021learning} as the default visual encoder if not stated.

More ablations are provided in the appendix, including effects of the finetuning set, different input settings, \etc.

\section{Conclusion}

In this work, we build a rigorous pipeline to collect high-quality and diverse video-music pairs, curating a comprehensive dataset \textbf{V2M}. Then, we propose VidMuse, a simple yet effective method for video-to-music generation. Our approach utilizes a Long-Short-Term approach to capture both local and global visual cues in the video, allowing for the generation of contextually rich and musically diverse outputs. To validate our method, we benchmark a series of state-of-the-art methods as baselines to compare with VidMuse. Through comprehensive quantitative studies and qualitative analyses, our method has demonstrated its superiority over the existing methods.

\textbf{Limitations}.
Our work achieves a significant advancement in video-to-music generation, but it still has some limitations. First, the current implementation relies on the EnCodec model~\cite{defossez2022highfi}, which sometimes exhibits a noticeable reconstruction loss for different genres of audio, potentially lowering the quality of the generated music. 
Second, our work solely focuses on video-to-music generation without exploring other input conditions. As such, our future work aims to overcome these limitations by integrating advanced codec technologies to enhance audio reconstruction fidelity, and exploring diverse controls during music generation.

\section{Acknowledgment}
The research was supported by NSFC (No. 62206234), the Early Career Scheme (ECS-Hong Kong University of Science and Technology 22201322), the Theme-based Research Scheme (T45-205/21-N), and the General Research Fund (No. 16212623) from the Hong Kong RGC, as well as the Generative AI Research and Development Center from InnoHK. We also thank Fortuna for discussions and Kyle for assisting in setting up the user study UI.

\clearpage
{
    \small
    \bibliographystyle{ieeenat_fullname}
    \bibliography{main}
}

\clearpage
\setcounter{table}{0}   
\setcounter{figure}{0}
\setcounter{page}{1}
\renewcommand{\thetable}{A\arabic{table}}
\renewcommand{\thefigure}{A\arabic{figure}}

\maketitlesupplementary

\begin{figure*}[t]
\centering
\includegraphics[width=0.8\linewidth]{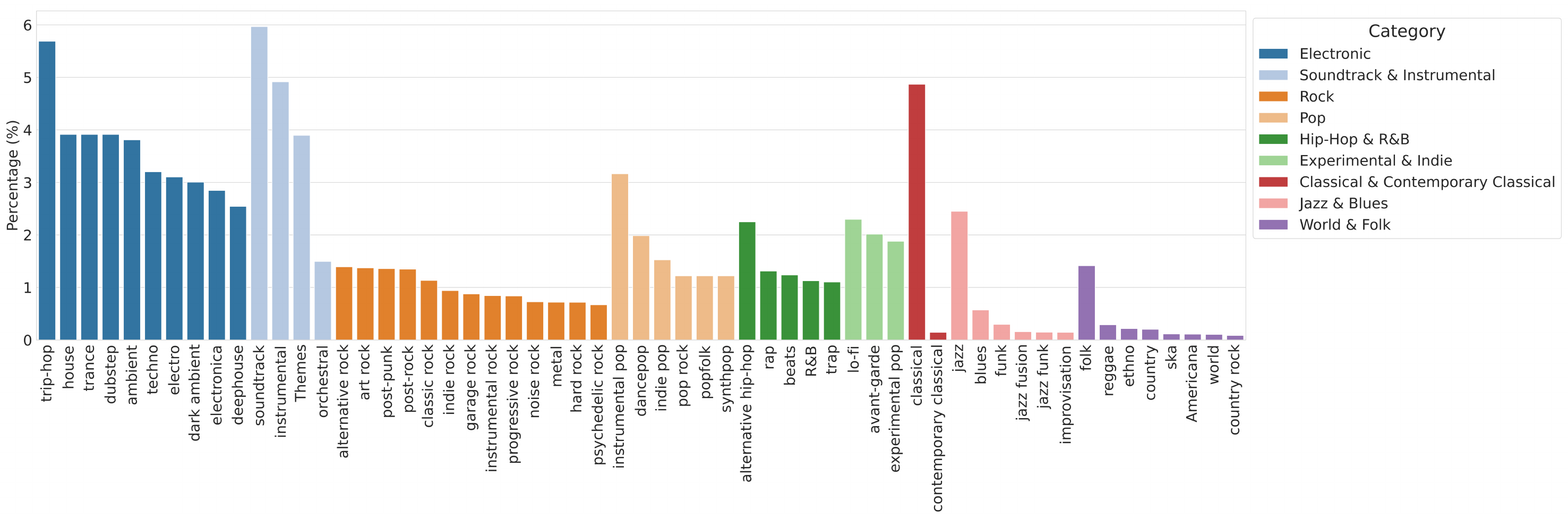}
\caption{Distribution of music genres in the dataset, showcasing the diverse representation of genres such as electronic, classical, and jazz.}
\label{fig:data_genre}
\end{figure*}

\begin{figure*}[t]
\centering
\includegraphics[width=0.8\linewidth]{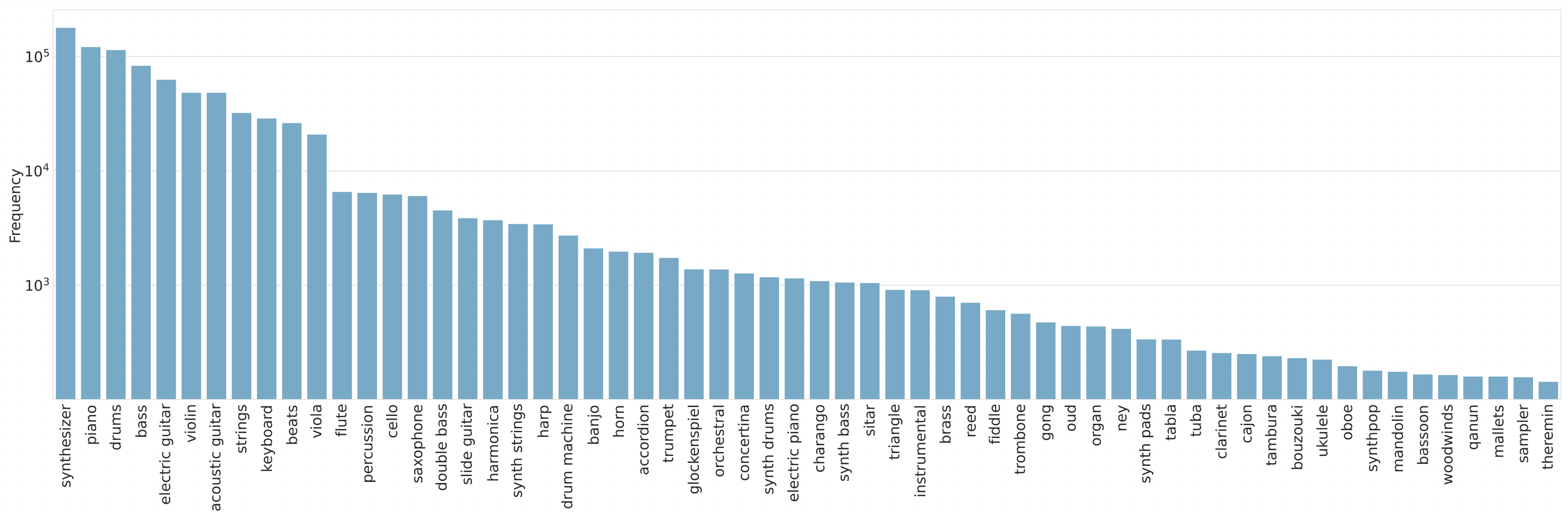}
\caption{Distribution of instruments in the dataset, emphasizing the frequent usage of synthesizers, pianos, and drums, while also including diverse instruments such as violins and saxophones.}
\label{fig:data_instrument}
\end{figure*}

\begin{figure*}[t]
\centering
\includegraphics[width=0.8\linewidth]{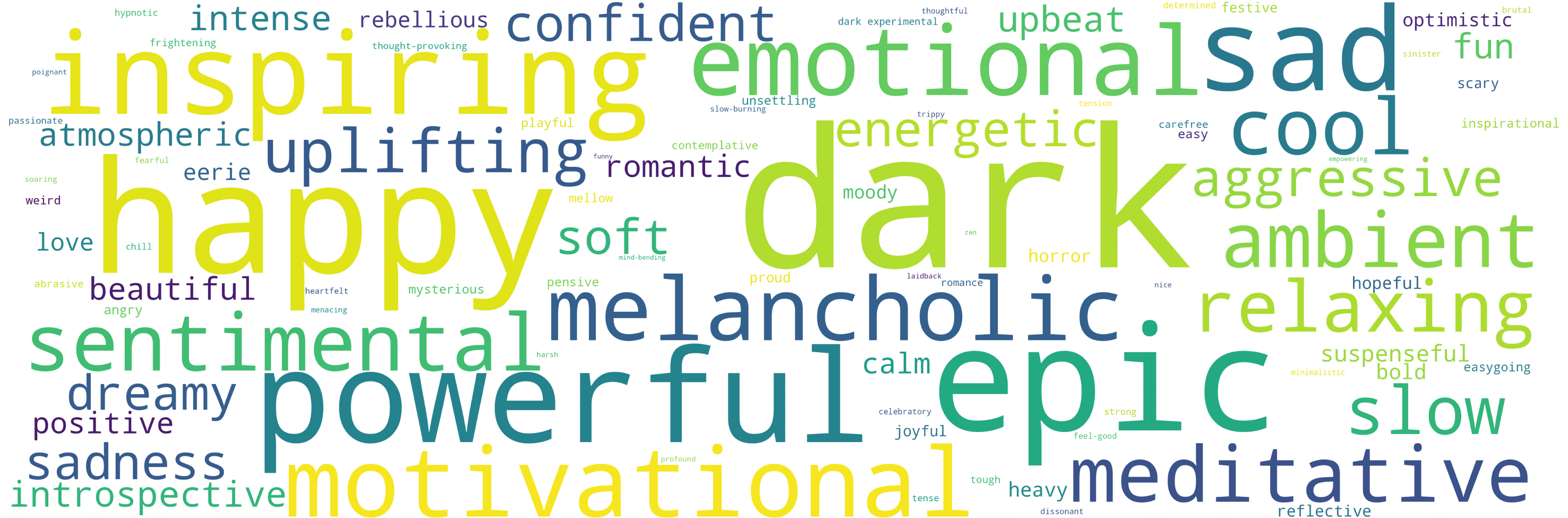}
\caption{Word cloud of mood labels in the dataset, highlighting the diversity of emotions such as inspiring, happy, powerful, and dark.}
\label{fig:data_word_cloud}
\end{figure*}

\section{Additional Experiments}
\label{sec:Additional_Experiments}
\begin{table*}[ht]
\caption{
\small{\textbf{Ablation studies on video duration and FPS.}} 
}
\label{tab:ablation_duation}
\footnotesize
\centering
\renewcommand{\arraystretch}{1.2}
\setlength{\tabcolsep}{2mm}
\begin{tabular}{c|c|c|c|c|c|c|c|c}
\Xhline{1.5pt}
\multicolumn{1}{c|}{\multirow{2}{*}{Duration(s)}} & \multicolumn{1}{c|}{\multirow{2}{*}{FPS}} & \multicolumn{7}{c}{Metrics} \\
\cline{3-9}
& & KL$~\color{red}{\downarrow}$& FD$~\color{red}{\downarrow}$ & FAD$~\color{red}{\downarrow}$ & density$~\color[RGB]{50, 200, 50}{\uparrow}$ & coverage$~\color[RGB]{50, 200, 50}{\uparrow}$ & Imagebind$~\color[RGB]{50, 200, 50}{\uparrow}$ & AR$~\color{red}{\downarrow}$\\
\Xhline{1pt}
5 & 2 & $0.820$ & $51.101$ & $4.117$ & $1.430$ & $0.74$ & $0.148$ & $7.00$\\
15 & 2 & $0.849$ & $41.131$ & $2.709$ & $1.406$ & $0.803$ & $0.181$& $5.33$\\
30 & 2& $0.843$ & $41.354$ & $\underline{2.413}$ & $1.487$ & $\underline{0.840}$ & $\textbf{0.193}$ & $3.67$\\ 
\hline
5 & 4 & $\textbf{0.800}$ & $51.540$ & $4.343$ & $1.271$ & $0.787$ & $0.145$& $7.17$\\
15 & 4 & $0.830$ & $41.154$ & $2.562$ & $1.278$ & $0.823$ & $0.176$& $5.17$\\
30 & 4 & $0.849$ & $\underline{40.032}$ & $2.418$ & $\underline{1.538}$ & $\textbf{0.843}$ & $\textbf{0.193}$ & $\underline{2.84}$\\ 
\hline
5 & 8 & $\underline{0.819}$ & $50.667$ & $4.069$ & $1.515$ & $ 0.743$ & $0.153$& $5.67$\\
15 & 8 & $0.857$ & $42.106$ & $2.790$ & $1.476$ & $0.753$ & $\underline{0.187}$& $6.00$\\
30 & 8 & $0.824$ & $\textbf{38.942}$ & $\textbf{2.299}$ & $\textbf{1.573}$ & $\textbf{0.843}$ & $0.180$ & $\textbf{2.17}$\\ 
\Xhline{1.5pt}
\end{tabular}
\end{table*}

Additional experiments focusing on model inputs, codebook patterns, and finetuning effects are provided in the appendix. These parts provide insight into the decision-making process for selecting the experimental configurations within the VidMuse framework.

\noindent\textbf{Exploration on model inputs.} 
To explore the impact of different video sampling rates and the duration of video segments in the Short-Term module on performance, we conducted ablation studies on input FPS and short-term segment duration, detailed in Table~\ref{tab:ablation_duation}. To intuitively assess the effectiveness of different settings, we employ an \textbf{Average Rank} (AR) metric. 
The AR metric ranks the results for a metric across all methods within the same table. The ranking result is from $1$ to $N$ (equals to the number of methods within the table), where $1$ is the best and $N$ is the worst. We eventually obtain AR results by averaging the ranking results for all metrics. Note that the AR results cannot be compared across different tables since this metric is designed to showcase the dominance of each method within one table clearly.
From Table~\ref{tab:ablation_duation}, we observe that increasing both FPS and duration tends to enhance model capabilities, suggesting that denser frame sampling yields a more detailed video representation, thereby improving music generation. Nevertheless, to balance computational costs and performance, we use a 30-second duration at 2 FPS as our optimal setting.

\begin{table*}[t!]
\caption{
\small{\textbf{Ablation studies on codebook pattern.}} 
}
\label{tab:ablation_codebook_pattern}
\footnotesize
\centering
\renewcommand{\arraystretch}{1.2}
\begin{tabular}{c|c|c|c|c|c|c}
\Xhline{1.5pt}
\multicolumn{1}{c|}{\multirow{2}{*}{Patterns}} & \multicolumn{6}{c}{Metrics} \\
\cline{2-7}
& KL$~\color{red}{\downarrow}$& FD$~\color{red}{\downarrow}$ & FAD$~\color{red}{\downarrow}$ & density$~\color[RGB]{50, 200, 50}{\uparrow}$ & coverage$~\color[RGB]{50, 200, 50}{\uparrow}$ & Imagebind$~\color[RGB]{50, 200, 50}{\uparrow}$ \\
\Xhline{1pt}
Parallel & $0.921$ & $68.603$ & $18.243$& $0.562$ & $0.183$ & $0.166$\\
Flatten & $\textbf{0.819}$ & $52.931$ & $4.260$ & $1.351$ & $0.500$ & $\textbf{0.201}$\\
Delay & $0.843$ & $\textbf{41.354}$ & $\textbf{2.413}$ & $\textbf{1.487}$ & $\textbf{0.840}$ & $0.193$ \\ 
Vall-E & $0.866$ & $57.286$ & $4.681$ & $1.148$ & $0.354$ & $0.189$ \\
\Xhline{1.5pt}
\end{tabular}
\end{table*}

\begin{table*}[!t]
\caption{
\small{\textbf{Ablation studies on the ratio of finetuning data.}} 
}
\label{tab:ablation_ft_data}
\footnotesize
\centering
\renewcommand{\arraystretch}{1.2}
\begin{tabular}{c|c|c|c|c|c|c}
\Xhline{1.5pt}
\multicolumn{1}{c|}{\multirow{2}{*}{\makecell[c]{Finetuning \\Data}}} & \multicolumn{5}{c}{Metrics} \\
\cline{2-7}
& KL$~\color{red}{\downarrow}$& FD$~\color{red}{\downarrow}$ & FAD$~\color{red}{\downarrow}$ & density$~\color[RGB]{50, 200, 50}{\uparrow}$ & coverage$~\color[RGB]{50, 200, 50}{\uparrow}$ & Imagebind$~\color[RGB]{50, 200, 50}{\uparrow}$ \\
 \Xhline{1pt}
0 & $\textbf{0.712}$ & $38.184$ & $3.956$ & $1.125$ & $0.583$ & $0.181$ \\
10k & $0.717$ & $34.667$ & $2.961$ & $0.856$ & $0.673$ & $0.196$ \\
20k & $0.734$ & $\textbf{29.946}$ & $\textbf{2.459}$ & $\textbf{1.250}$ & $\textbf{0.730}$ & $\textbf{0.202}$ \\
40k & $0.776$ & $41.075$ & $3.557$ & $1.094$ & $0.726$ & $0.195$ \\
60k & $0.828$ & $40.160$ & $2.844$ & $0.977$ & $0.660$ & $0.192$ \\
 
\Xhline{1.5pt}
\end{tabular}
\end{table*}

\noindent\textbf{Codebook Pattern.} The exploration of codebook interleaving patterns has attracted attention from researchers across several domains~\cite{zeghidour2021soundstream, wang2023neural, copet2024simple, yang2023hifi, lan2023stack}. In our ablation study focusing on the patterns, we find that while the Parallel and Vall-E~\cite{wang2023neural} patterns align with the findings for text-to-music generation in MusicGen~\cite{copet2024simple}, the flattened codebook pattern does not consistently exceed the performance of the delay pattern in tasks of generating music from video. The delay pattern, notable for its relatively low computational cost, is therefore selected for our implementation. The results of this study are presented in \cref{tab:ablation_codebook_pattern}.

\noindent\textbf{Finetuning Effect.} Our ablation study on the effects of the data scale during finetuning, as detailed in Table~\ref{tab:ablation_ft_data}, highlights a balance between data size and model performance. Despite not performing best in all the metrics, the model finetuned with 20k pair data emerges as our choice. The 20k data offers a compelling trade-off: it significantly improves performance across key metrics without requiring the extensive computational resources that larger datasets demand.
The results also validate the effectiveness of our ranking strategy based on ImageBind-AV scores (detailed in Appendix~\ref{sec:Details_of_Dataset_Construction}), showing that prioritizing videos with higher audio-visual alignment improves finetuning data quality and enhances model performance.

\section{Details of Dataset Construction}
\label{sec:Details_of_Dataset_Construction}
\noindent{\textbf{Coarse Filtering.}} We design a rule-based filtering strategy for initial data screening. First, we perform illegal video and audio filters, which filter out the video without an audio track or a video track. Next, we apply a duration filter to filter out videos based on their duration, excluding those that are either too long (over 480 seconds) or too short (under 30 seconds). Additionally, we implement a domain filter to examine metadata and exclude specific categories such as \textit{Interview}, \textit{News}, and \textit{Gaming}, which often have background music that lacks semantic alignment with the visual content. We also filter out videos containing inappropriate content, such as violence or explicit material.

\noindent{\textbf{Fine-grained Filtering.}} To further ensure the quality of our data, we conducted additional audio and visual analyses.
For the audio analysis, raw videos may contain audio segments without music, such as speech, silence, \etc. To ensure the final dataset consists of high-quality video-music pairs, we retain only those videos with a larger portion of music content. We utilize the sound event detection model PANNs~\cite{kong2020panns}, which provides frame-level event labels across the entire video to identify music events. Based on the observation from a subset of videos, we define two thresholds, \ie, a confidence threshold and a duration threshold, for analyzing the music event.
The confidence threshold is set at \(0.5\), indicating an audio frame is considered a music event if the PANNs model predicts the probability of the ``Music'' label to be over \(0.5\). The duration threshold of a music event requires that at least \(50\%\) of the audio's frames are classified as music events for the video to be considered valid.

For the visual analysis, some videos only consisting of static images will be removed. Specifically, we uniformly sample multiple temporal windows without overlap from the video. Within each window, we use Structural Similarity Index Measure (SSIM)~\cite{wang2004image} between the first frame and the last frame. By aggregating average SSIM values from all temporal windows, we remove the videos with average SSIM values lower than a threshold of 0.8, empirically.

\noindent{\textbf{Music Source Separation.}} Since the irrelevant human speech in videos poses a negative impact on music generation, we apply music source separation to process the videos. We employ Demucs~\cite{rouard2023hybrid} as the music source separation model to filter out the speech signals.

\noindent{\textbf{Audio-Video Alignment Ranking.}} ImageBind-AV~\cite{girdhar2023imagebind} scores usually reflect the semantic correlation between the vision and audio modality. To construct a high-quality subset with better alignment, we compute the ImageBind-AV scores for all the data and rank them accordingly.

After filtering and ranking, we split the final videos into the training set, \textit{V2M}, from all the paired data. The top 20K pairs are selected to form the finetuning subset, \textit{V2M-20K}. In addition, we randomly sample 1,000 videos excluded from the training set. These 1,000 videos are then further evaluated by five human experts based on audio quality and the degree of audio-visual alignment. Ultimately, the top 300 high-quality videos are selected as a test set, termed as \textit{V2M-bench}.

\section{Additional Dataset Analysis}
\noindent\textbf{Music Genre Distribution.} 
To better understand the diversity of our dataset, we analyze the distribution of music genres across all selected video-music pairs. The results are illustrated in Fig.~\ref{fig:data_genre}. As shown, the dataset covers a wide range of genres, including but not limited to electronic, classical, pop, and rock. The diversity in genres ensures that the dataset provides a comprehensive foundation for the task of video-to-music generation, enabling robust performance across various musical styles.

\noindent\textbf{Instrument Usage Distribution.} 
We also analyze the usage of different instruments within the dataset. The distribution is shown in Fig.~\ref{fig:data_instrument}. The frequent occurrences of synthesizers, pianos, and drums, along with a variety of other instruments, ensure the ability to capture diverse musical elements in the video-to-music generation task.

\noindent\textbf{Mood Information.} 
In addition to genres and instruments, we also explore the mood information present in the music data. A word cloud representation of the mood labels is shown in Fig.~\ref{fig:data_word_cloud}, where the font size corresponds to the frequency of each mood label. Commonly occurring moods include \textit{inspiring}, \textit{happy}, \textit{dark}, \textit{powerful}, and \textit{sentimental}, showcasing the emotional diversity of the dataset. This emotional richness enhances the dataset's capacity to generate music that aligns closely with the mood conveyed in videos.

All music-related metadata, including genre, instrument, and mood, is annotated using Qwen2-Audio, a state-of-the-art (SOTA) model for music understanding.

\section{Details of Evaluation Metrics}
\label{sec:Details_of_Evaluation_Metrics}

\noindent\textbf{Frechet Audio Distance (FAD)} is a reference-free evaluation metric for assessing audio quality. Similar to Frechet Image Distance (FID)\cite{heusel2017gans}, it compares the embedding statistics of the generated audio clip with ground truth audio clips. A shorter distance usually denotes better human-perceived acoustic-level audio quality. However, this metric cannot reflect semantic-level information in audio. We report the FAD based on the VGGish\cite{hershey2017cnn} feature extractor.

\noindent\textbf{Frechet Distance (FD)} measures the similarity between generated samples and target samples in audio generation fields. It's similar to FAD but uses a PANNs feature extractor instead. PANNs\cite{kong2020panns} have been pre-trained on AudioSet\cite{gemmeke2017audio}, one of the largest audio understanding datasets, thus resulting in a more robust metric than FAD.

\noindent\textbf{Kullback-Leibler Divergence (KL)} reflects the acoustic similarity between the generated and reference samples to a certain extent. It is computed over PANNs' multi-label class predictions.

\noindent\textbf{Density and Coverage}~\cite{naeem2020reliable} measures the fidelity and diversity aspects of the generated samples. 
Fidelity measures how closely the generated samples match the real ones, while diversity assesses whether the generated samples capture the full range of variation found in real samples. We use CLAP\cite{wu2023large} embeddings for manifold estimation. 

\noindent\textbf{Imagebind Score}~\cite{girdhar2023imagebind} assesses to what extent the generated music aligns with the videos. Despite the fact that Imagebind extends the CLIP to six modalities, we only use the branches of audio and vision.  
Since we use ImageBind to filter out video-audio pairs with a low matching score during dataset construction, the ImageBind score is naturally used in our evaluation. We acknowledge that ImageBind is not specifically trained on music data, which may limit its effectiveness in capturing the full complexity of video-music alignment. However, it remains the most suitable option available for this task at present.

\section{Details of the Inference Process}
\label{sec:Details_of_Inference_Process}
When predicting music on videos of arbitrary length, maintaining music consistency and coherence is particularly important.
However, it leads to a significant challenge on computational resources due to the quadratic dependency of transformers-based models on sequence length~\cite{zaheer2020big,beltagy2020longformer}. 

To address this problem, we adopt a sliding window approach for inferring the whole video.
During inference, given an input video with a length of $ L $, we define $ L_s $ as the length of the sliding window and $ O $ as the overlap between consecutive windows. With the window start position $ t $ initially set to $ 0 $, the inference involves the following steps compactly while $ t + L_s \leq L $: (1) using a visual encoder to extract feature representations $ \textbf{X} $ and capture long-term dependencies $\textbf{X}_{l} $; (2) collecting embeddings within the window $[t, t+L_s]$ to obtain $\textbf{X}_{s} $; (3) predicting the music tokens $ \bar{\textbf{Y}} $ for the reduced window $[t, t+L_s-O]$ based on $\textbf{X}_{l} $ and $\textbf{X}_{s} $; (4) decoding $ \bar{\textbf{Y}} $ to the predicted audio $ \bar{\textbf{A}} $ using the audio decoder; (5) move the window forward by setting $ t = t + L_s - O $, and repeating steps (2) to (5) until the end of the video.

After finishing the above steps, we can concatenate all musical segments to form a cohesive piece of music that aligns in duration with the video.


\section{Qualitative Analysis}

\begin{figure*}[t]
\centering
\includegraphics[width=0.8\linewidth]{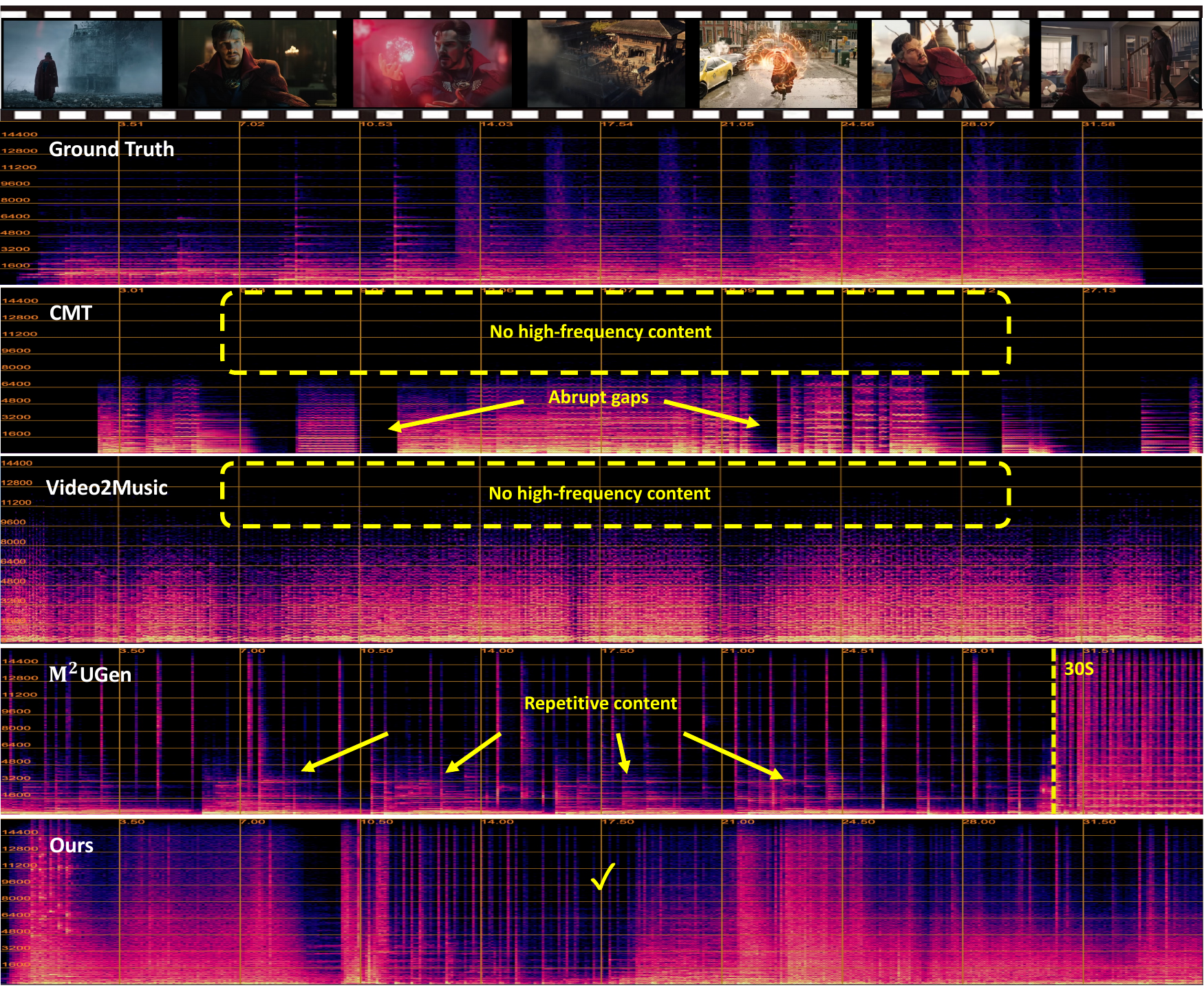}
\caption{Qualitative Comparison results on sound spectrograms produced by different methods.}
\label{fig:qualitative_result}
\end{figure*}

In Fig.~\ref{fig:qualitative_result}, our qualitative analysis highlights specific limitations of CMT, Video2Music, and M$^2$UGen. 
CMT and Video2Music extract visual cues to generate symbolic music, \ie, MIDI notes. However, CMT's training strategy for symbolic music generation leads to discontinuities, particularly for slowly changing or static frames, where the model fails to predict symbolic music notes, resulting in periods of silence. Additionally, the approach of predicting MIDI notes and then rendering them into audio, as employed by both CMT and Video2Music, lacks high-frequency content, negatively affecting auditory perception.
M$^2$UGen utilizes LLMs to fuse multimodal representation and then project LLMs' embeddings into music via a text-to-music generation model. 
However, this approach relies on text embeddings as intermediaries, which causes the loss of visual information and restricts the model's ability to detect nuanced visual variations. As a result, the music generated by this method usually showcases repetitive musical themes and suffers from a lack of diversity, as evidenced in Fig.~\ref{fig:qualitative_result} and the supplementary videos.
The last row of Fig.~\ref{fig:qualitative_result} demonstrates that our Long-Short-Term (LST) approach is capable of generating music that is rich in diversity and semantically consistent with the video.

\section{User Study Interface}
\begin{figure*}[t]
\centering
\includegraphics[width=0.8\linewidth]{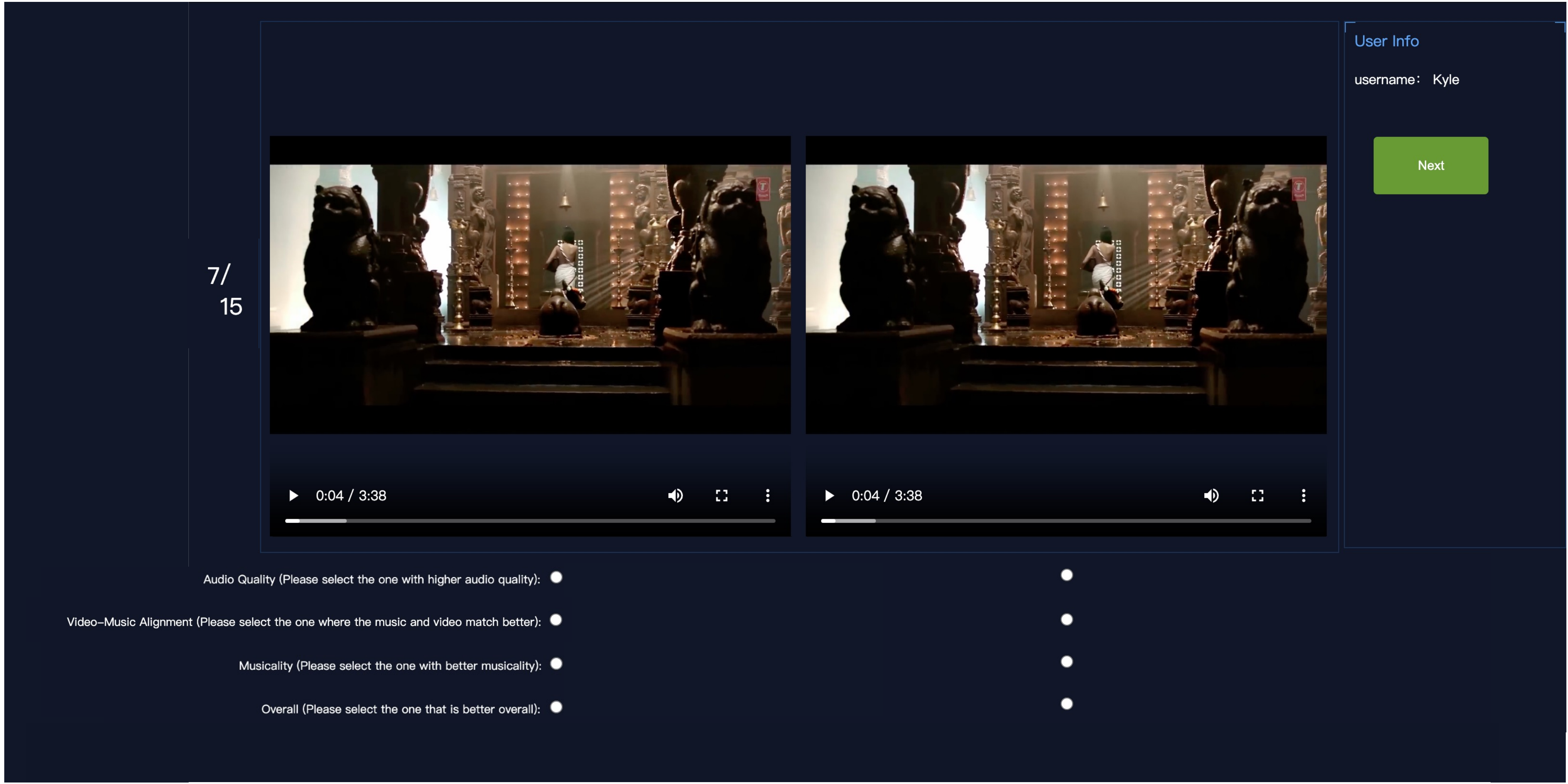}
\caption{User study process. Participants evaluate the videos based on four criteria: Audio Quality, Video-Music Alignment, Musicality, and Overall Assessment.}
\label{fig:user_study_UI}
\end{figure*}

Fig.~\ref{fig:user_study_UI} illustrates the A/B test interface used during the user study. Participants evaluated the videos based on four criteria: Audio Quality, Video-Music Alignment, Musicality, and Overall Assessment. 
This interface shows participants comparing two videos side-by-side and selecting the better one for each criterion.

\section{Supplementary Videos}  
For additional insights and demonstrations, we kindly refer readers to our supplementary video for a comprehensive showcase of our method's performance.

\end{document}